Purdue University

# Trajectory tracking control of a Remotely Operated Underwater Vehicle based on Fuzzy Disturbance Adaptation and Controller Parameter Optimization

A ME697 (Intelligent systems) Final Project Report

Yang, Hanzhi

4-26-2024

# Table of Contents

# Abstract

The exploration of under-ice environments presents unique challenges due to limited access for scientific research. This report investigates the potential of deploying a fully actuated Remotely Operated Vehicle (ROV) for shallow area exploration beneath ice sheets. Leveraging advancements in marine robotics technology, ROVs offer a promising solution for extending human presence into remote underwater locations. To enable successful under-ice exploration, the ROV must follow precise trajectories for effective localization signal reception. This study develops a multi-input-multi-output (MIMO) nonlinear system controller, incorporating a Lyapunov-based stability guarantee and an adaptation law to mitigate unknown environmental disturbances. Fuzzy logic is employed to dynamically adjust adaptation rates, enhancing performance in highly nonlinear ROV dynamic systems. Additionally, a Particle Swarm Optimization (PSO) algorithm automates the tuning of controller parameters for optimal trajectory tracking. The report details the ROV dynamic model, the proposed control framework, and the PSO-based tuning process. Simulation-based experiments validate the efficacy of the methodology, with experimental results demonstrating superior trajectory tracking performance compared to baseline controllers. This work contributes to the advancement of under-ice exploration capabilities and sets the stage for future research in marine robotics and autonomous underwater systems.

# I. Introduction

## i. Background

The Earth's oceans, spanning 71% of its surface, encounter a unique challenge in the 12% of areas covered by ice, limiting accessibility for scientific research. Few methods presently exist for routine benthic survey and sampling operations under ice in high latitudes. In contrast, present day blue-water oceanographic methods for survey and sampling are extensive, including ship-based sensing, lowered instruments such as dredges, tethered Remotely Operated Vehicles (ROVs) and untethered vehicles such as Human Occupied Vehicles (HOVs), Autonomous Underwater Vehicles (AUVs), and Hybrid Remotely Operated Vehicles (HROVs). Only a few of these methods, principally lowered instruments deployed from icebreaking ships (H. Edmonds 2003), or from ice camps, e.g. (W. M. Smethie 2011), are regularly practiced for under-ice sampling and survey operations.

In this project, we examined the potential of deploying a fully actuated ROV for under-ice shallow area exploration. With the advances of marine robotics technology, ROVs have been widely applied in scientific and industrial fields. The ROVs extend the actions and intelligence of humans to remote locations. An example of an ROV model is shown in Fig. I-1. In practical applications, light-weight-observation ROVs have become a powerful tool for offshore observation due to their advantages of low-cost manufacture, reliable performance, and flexible usage.

To deploy an ROV in an under-ice water region, the vehicle needs to follow certain trajectories for better localization signals. As mentioned in (Alexander 2012), the underwater localization signal intensities under ice sheets are greatly affected by depth and patterns of the localization sensors. Therefore, to accomplish a successful under-ice water exploration with continuous communication between the vehicle and over-ice main control station, it is crucial for the ROV to stay in the pre-planned course, which requires a well-designed control system for trajectory tracking.

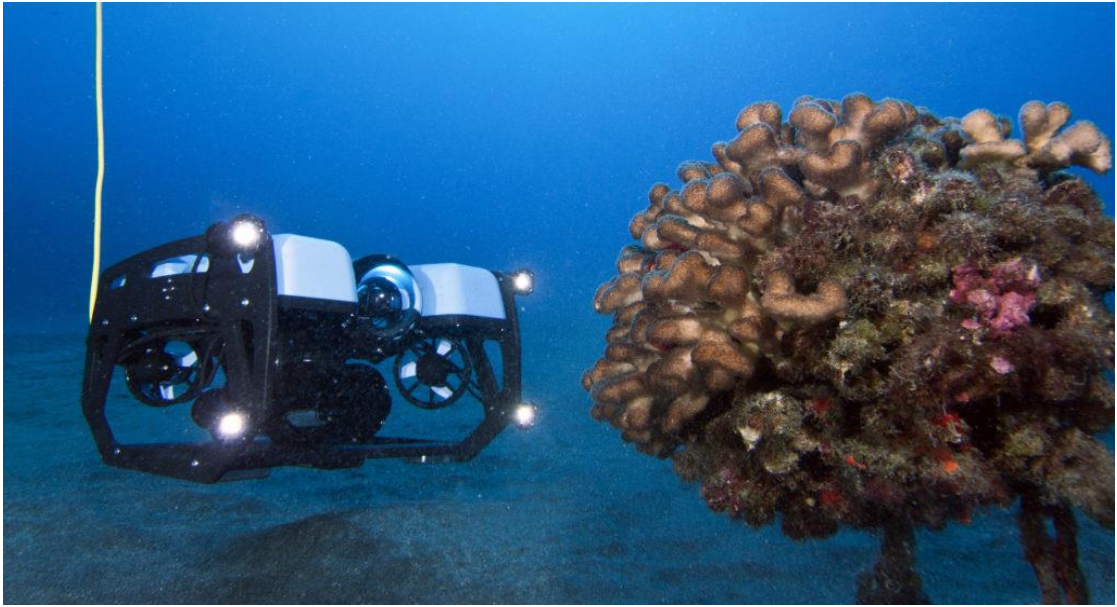

*Figure I-1. A BlueROV2 operating in the sea (BlueRobotics 2024).*

## ii.    Related works

In addressing the ROV tracking challenge, numerous control strategies have been devised. In an early work, a tracking controller that combines proportional-derivative (PD) control and adaptive algorithm is proposed in (Kreuzer 2007). In (J. Cervantes 2016), backstepping and feedback linearization techniques are applied to design the tracking controller. The energy-efficient issue is considered in (M. Zhang 2017), and then a region tracking controller is proposed to drive the vehicle to the desired target point. A self-tuning tracking controller is developed in (Subudhi 2017), and a Lyapunov-based model predictive control (LMPC) scheme is presented in (C.

Shen 2018). In (Whitcomb 2018), two model-based tracking controllers are designed to enable ROV to perform the tracking task. An adaptive high order sliding mode controller for trajectory tracking is developed in (J. Guerrero 2019). In (C. Z. Ferreira 2018), a multivariable control strategy is presented to achieve the tracking task.

### iii. Problem statement

In this work, to deploy ROVs in under-ice waters, we firstly designed a multi-input-multi-output (MIMO) nonlinear system controller that guarantees system stability based on its Lyapunov function. To lower the influence of unknown environmental disturbances, we added an adaptation law to the proposed control system such that the actual disturbances can be estimated. To further improve the performance of the proposed linear adaptation law on a highly nonlinear ROV dynamic system, we applied fuzzy logic to vary the adaptation rates based on the output errors. Due to the large number of controller parameters, manually tuning can be difficult. Therefore, we utilized particle swarm optimization (PSO) algorithm to find the best combination of controller parameters to gain the lowest optimal control cost function value.

This report is set up as follows: Chapter 2 analyzes the ROV dynamic model, presents the proposed control law and fuzzy disturbance adaptation law, and introduces the PSO-based controller tuning process structure; Chapter 3 validates the proposed methodology through a simulation-based experiment; Chapter 4 shows the experimental results of controller tuning and compare the trajectory tracking performance by the proposed control method to that by the baseline controller; Chapter 5 concludes the report and considers the future works.

## II. Methodology

In this chapter, we analyze the dynamic model of a typical ROV system and then design a nonlinear controller based on backstepping and feedback linearization methodologies. To lower the influence of environmental disturbances, which is unknown to the ROV system, an adaptation law is added, and the adaptation rates are determined by a fuzzy inference system (FIS). The overall control and adaptation system is proved to be stable by Lyapunov functions. The controller parameters are tuned based on particle swarm optimization (PSO) method.

### i. System model

A general dynamic model of ROVs can be described in a body-fixed coordinate frame and a global coordinate frame as shown in Fig. II-1. The kinematic equation of ROVs is given by

$$\dot{\eta} = J(\eta)v \tag{2.1}$$

in which $J(\eta)$ is the coordinate transformation matrix that yields

$$J(\eta) = \begin{bmatrix} J_t(\eta) & \emptyset \\ \emptyset & J_r(\eta) \end{bmatrix} \tag{2.2}$$

and the translational and rotational transformation matrix $J_t(\eta)$ and $J_r(\eta)$ are expressed as

$$J_t(\eta) = \begin{bmatrix} c\psi c\theta & -s\psi c\phi + c\psi s\theta s\phi & s\psi s\theta + c\psi c\phi s\theta \\ s\psi c\theta & c\psi s\theta + s\phi s\theta s\psi & -c\psi s\theta + s\theta s\psi c\phi \\ -s\theta & c\theta s\phi & c\theta s\phi \end{bmatrix} \tag{2.3}$$

$$J_r(\eta) = \begin{bmatrix} 1 & s\phi t\theta & c\phi t\theta \\ 0 & c\phi & -s\phi \\ 0 & \dfrac{s\phi}{c\theta} & \dfrac{c\phi}{c\theta} \end{bmatrix} \tag{2.4}$$

in which $s(\cdot) = \sin(\cdot)$, $c(\cdot) = \cos(\cdot)$, $t(\cdot) = \tan(\cdot)$.

*Assumption 1:* The roll and pitch angles of the ROV models analyzed in this work are limited, i.e.

$$|\phi| < \frac{\pi}{2} \tag{2.5}$$

$$|\theta| < \frac{\pi}{2} \tag{2.6}$$

As a result of the assumption, $\det(J(\eta)) \neq 0$, and $J_r(\eta)$ is positive definite.

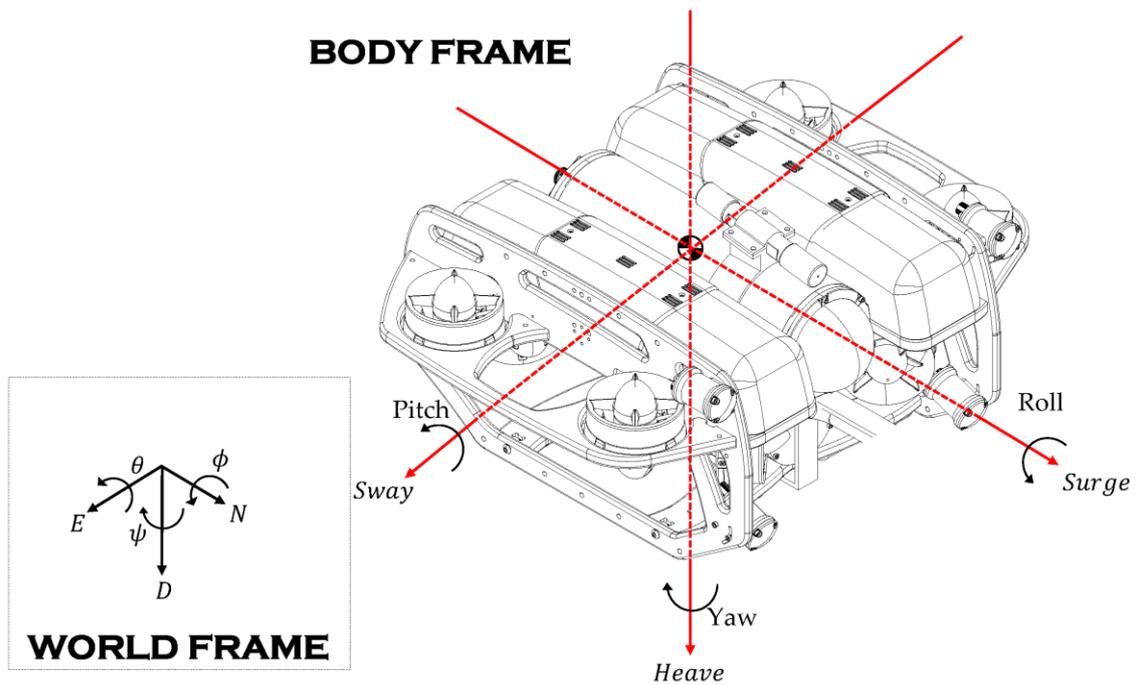

*Figure II-1. Global and body reference frames on an ROV model.*

The 6 degrees of freedom (DOF) equations of motion (EoM) considering unknown environmental disturbances in the body frame are expressed in matrix form as

$$M\dot{v} + C(v)v + D(v)v + g(\eta) = \tau + \tau_d \tag{2.7}$$

where $v = [u, v, w, p, q, r]^T$ is the body frame velocity vector, $\eta = [X, Y, Z, \phi, \theta, \psi]^T$ is the global frame position and orientation vector, $M$ is the inertia matrix including the added mass, i.e. $M = M_{rov} + M_a$, $C(v)$ is the Coriolis matrix and centrifugal terms including the added mass, i.e. $C(v) = C_{rov}(v) + C_a(v)$, $D(v)$ is the damping matrix, and $g(\eta)$ is a vector including gravitational and buoyancy forces and moments. The detailed elements in these matrices are shown in Appendix I.

*Property 1:* The parameter matrices in Eq. 2.7 have the following properties:

$$M = M^T > 0 \tag{2.8}$$

$$C(v) = -C^T(v), \forall v \in \mathbb{R}^{6 \times 1} \tag{2.9}$$

$$D(v) > 0, \forall v \in \mathbb{R}^{6 \times 1} \tag{2.10}$$

$\tau_d$ is the unknown environmental disturbance vector. For ROV operations, such disturbances are usually produced by water currents and waves. Considering the under-ice deployment scenarios, the following assumption is made to simplify the problem,

*Assumption 2:* The disturbance vector $\tau_d$ is bounded and of low frequency, i.e.

$$|\tau_d|_i \leq (d_{max})_i \tag{2.11}$$

$$|\dot{\tau}_d|_i \leq (\dot{d}_{max})_i \tag{2.12}$$

in which $d_{max} = [d_u, d_v, d_w, d_p, d_q, d_r]^T > 0$.

$\tau$ is the control force and torque vector. Generally, on an ROV model, there is a low-level open-loop controller that calculates the actual command inputs for the onboard thrusters to achieve the target $\tau$. In this work, for simplicity, $\tau$ is considered to be the control input which is determined by the proposed control system.

## ii. Control law

An ideal multi-input-multi-output (MIMO) controller is designed based on backstepping and feedback linearization methods to guarantee the asymptotically stability of the closed-loop system. Given a reference trajectory $\eta_d \in \mathbb{R}^{6 \times 1}$, the tracking error is defined as

$$e = \eta - \eta_d \tag{2.13}$$

and a sliding surface is defined as

$$s = \dot{e} + k_1 e \tag{2.14}$$

where $k_1 = diag(k_{11}, k_{12}, k_{13}, k_{14}, k_{15}, k_{16}) \in \mathbb{R}^{6 \times 6} > 0$ is a set of controller parameters. Taking the derivative of Eq. 2.1 gives

$$\ddot{\eta} = \underbrace{\dot{J}v}_{A} + J\dot{v} \tag{2.15}$$

In the following equations, $\dot{J}v$ is represented by $A$ for clearer indications. Substituting Eq. 2.7 into Eq. 2.15 yields an Input/Output (I/O) form of the system model expressed as

$$\ddot{\eta} = A + JM^{-1}(-(C+D)v - g + \tau_d) + JM^{-1}\tau \tag{2.16}$$

Now a control law to track the reference trajectory is designed as

$$\tau = (JM^{-1})^T(-k_1\dot{e} + \ddot{\eta}_d + \left(JM^{-1}((C+D)v + g - \widehat{\tau_d})\right) - A - k_2 s) \tag{2.17}$$

where $k_2 = diag(k_{21}, k_{22}, k_{23}, k_{24}, k_{25}, k_{26}) \in \mathbb{R}^{6 \times 6} > 0$ is another set of controller parameters and $\widehat{\tau_d}$ is the estimated value of the unknown bounded environmental disturbance.

*Lemma 1:* If the estimated disturbance tracks the actual values well, i.e. $\widehat{\tau_d} = \tau_d$, the control law in Eq. 2.17 is able to stabilize the system in Eq. 2.1 and Eq. 2.7 to track the reference trajectory $\eta_d$.

*Proof:* Consider the following Lyapunov function candidate

$$V_c = \frac{1}{2}\sum_{i=1}^{6} s_i^2 = \frac{1}{2}s^T s \qquad (2.18)$$

Its derivative is

$$\dot{V}_c = s^T \dot{s} = s^T(k_1 \dot{e} - \ddot{\eta}_d - (JM^{-1}((C+D)v + g - \tau_d) + A + JM^{-1}\tau \qquad (2.19)$$

And substituting in the control law in Eq. 2.17 yields

$$\dot{V}_c = s^T(JM^{-1}\tau_d - JM^{-1}\widehat{\tau_d} - k_2 s) \qquad (2.20)$$

By the assumption made in Lemma 1 that $\widehat{\tau_d} = \tau_d$,

$$\dot{V}_c = -s^T k_2 s \leq 0 \qquad (2.21)$$

which indicates that $s$ is converging to zero. By Eq. 2.14, since $k_1$ is positive defined, $e$ is also converging to zero. Hence, the control law enables the ROV system to track the reference trajectory.

### iii. Disturbance adaptation law

A linear disturbance adaptation law is designed based on backstepping method and Lyapunov functions. Given the assumption that disturbance is expected to be bounded and of low frequency, its adaptation law is designed to be

$$\dot{\widehat{\tau_d}} = \Gamma(M^{-1}J)^T s \qquad (2.22)$$

where $\Gamma = diag(\gamma_1, \gamma_2, \gamma_3, \gamma_4, \gamma_5, \gamma_6) \in \mathbb{R}^{6\times 6} > 0$ is the adaptation rate coefficient.

*Lemma 2:* The adaptive control system given by Eqs. 2.17 and 2.22 guarantees the ROV system's asymptotically stability.

*Proof:* Define the disturbance estimation error

$$\widetilde{\tau_d} = \widehat{\tau_d} - \tau_d \tag{2.23}$$

and consider an adaptation Lyapunov function candidate

$$V_a = \frac{1}{2}\widetilde{\tau_d}^T \Gamma^{-1} \widetilde{\tau_d} \tag{2.24}$$

Adding the controller Lyapunov function in Eq. 2.18 yields the entire adaptive control system Lyapunov function

$$V = V_c + V_a = \frac{1}{2}s^T s + \frac{1}{2}\widetilde{\tau_d}^T \Gamma^{-1} \widetilde{\tau_d} \tag{2.25}$$

the derivative of which, after substituting the control law in Eq. 2.17, is

$$\dot{V} = s^T(-JM^{-1}\widetilde{\tau_d} - k_2 s) + \dot{\widehat{\tau_d}}^T \Gamma^{-1} \widetilde{\tau_d} \tag{2.26}$$

Assumption 2 indicates that

$$\dot{\widetilde{\tau_d}} = \dot{\widehat{\tau_d}} \tag{2.27}$$

Hence, substituting the adaptation law in Eq. 2.22 to Eq. 2.26 gives

$$\dot{V} = -s^T k_2 s + \left(-s^T JM^{-1} + \dot{\widehat{\tau_d}}^T \Gamma^{-1}\right)\widetilde{\tau_d} = -s^T k_2 s \leq 0 \tag{2.28}$$

which proves the stability of the proposed adaptive control system such that the disturbance estimate error converges to zero.

### iv. Fuzzy adaptation rate

One issue with implementing the proposed adaptive control system to the ROV system is that Lemma 1 and Lemma 2 only proves stability in infinite time, which means that without careful tuning, it is possible that in finite time the system could not reach stability. In addition, the ROV system is highly nonlinear, so there are limitations to apply a linear adaptation law to it. One example of a failed disturbance estimation in finite time is shown in Fig. IV-5 in Chapter 4. It is difficult to tune the adaptive control system to accomplish fine estimation convergence with constant adaptation rates. So, in this work, a type-1 Mamdani FIS is introduced to vary the adaptation rates based on output errors. The basic idea is to overcome the nonlinearity of the system by a series of linear adaptation laws instead of only one.

Mamdani systems are composed of IF-THEN rules of the form "IF $X$ is $A$ THEN $Y$ is $B$", in which the IF part "$X$ is $A$" is called the antecedent of the rule, and the THEN part "$Y$ is $B$" is called the consequent of the rule (Izquierdo 2018). The defuzzification step transforms the aggregated fuzzy set $\mu_{Mamdani|x}$ into one single crisp number. Standard Mamdani systems use the Center of Gravity (COG) defuzzification method. This method returns the projection (on the horizontal axis) of the center of gravity of the area under the membership function $\mu_{Mamdani|x}$.

In this work, one Mamdani FIS with a designed number of rules is applied to each DOF in body frame dynamics. The rules consider the tracking errors after being transformed to body frame dynamics and is given by

$$IF\ |JM^{-1}s|_i = A_{ij}\ THEN\ \gamma_i = B_{ij} \tag{2.29}$$

where $i = 1, \dots, 6$ represents each DOF of ROV in its body frame and $j = 1, \dots, \#\ of\ rules$ represents each rule set for the corresponding antecedent conditions. The fuzzy rules of adaptive rate coefficients make it easier for the adaptation law to vary the adaptation rates based on the output errors and thus overcome the nonlinearity of the ROV system in order to gain disturbance estimation convergence in finite time.

### v. PSO-based controller tuning

Since the proposed controller utilizes 12 controller parameters, it is difficult to manually tune the controller. Hence, in this work, PSO algorithm is applied to find the optimal combination of controller parameters to gain best tracking performance.

PSO algorithm is an optimization algorithm abstracted by simulating the fight and foraging behavior of birds in nature (Ou 2006). In a typical PSO set up, a fixed number particle is involved. Each particle is associated with a vector of fixed number of elements. To start with, these elements are randomly initialized and the whole system of particles enters an iterative process. At the end of each iteration, the performance of each particle is calculated in terms of the closeness to the objective function by substituting the values of the elements of the vector in the objective function. At the end of every iteration, it may so happen that one among of the particles comes out with the best results. That particle is referred to as the globally best particle of the iteration. In the next iteration, another particle may become the globally best particle.

Besides, each particle over a number of iterations may exhibit different performances in each iteration and by considering the past and the present iteration, the certain combination of the elements of the vector for a particular iteration might stand the best performing vector pertaining to that particular particle. This particular vector is the personal best of the said particle. All the particles are updated with the best vector that gives the best results.

Based on the vector of the personal best of each particle and based on the vector of the globally best performing particle, a velocity is estimated, and this velocity is added up respectively with the updated best performing vector of each particle. Thus, at the end of each iteration, the vectors of all the particles are updated and the performance evaluated with

$$v_i(t+1) = wv_i(t) + c_1 r_1 \left(P_{best_i} - x_i(t)\right) + c_2 r_2 (G_{best_i} - x_i(t)) \tag{2.30}$$

$$x_i(t+1) = x_i(t) + v_i(t+1) \tag{2.31}$$

in which $P_{best}$ is the personal best vector, $G_{best}$ is the global best vector, $x$ is the position of the particles, $v$ is the velocity of the particles, $r_1$ and $r_2$ are random numbers, and $w$, $c_1$, $c_2$ are optimizer design parameters.

In the case of tuning the proposed controller, a cost function is defined as

$$J = \int_0^{t_f} s^T Q s + u^T R u \; dt \tag{2.32}$$

in which $Q \in \mathbb{R}^{6 \times 6} > 0$ and $R \in \mathbb{R}^{6 \times 6} > 0$ are design parameters and $t_f$ is the total time to run the simulation. By taking the positions of each particle as the controller parameters, it is expected that the optimal combination of the 12 controller parameters is given by

$$\{k_1, k_2\} = argmin_{\{k_1 > 0, k_2 > 0\}} J \tag{2.33}$$

The PSO-based controller tuning process structure in this work is shown in Fig. II-2.

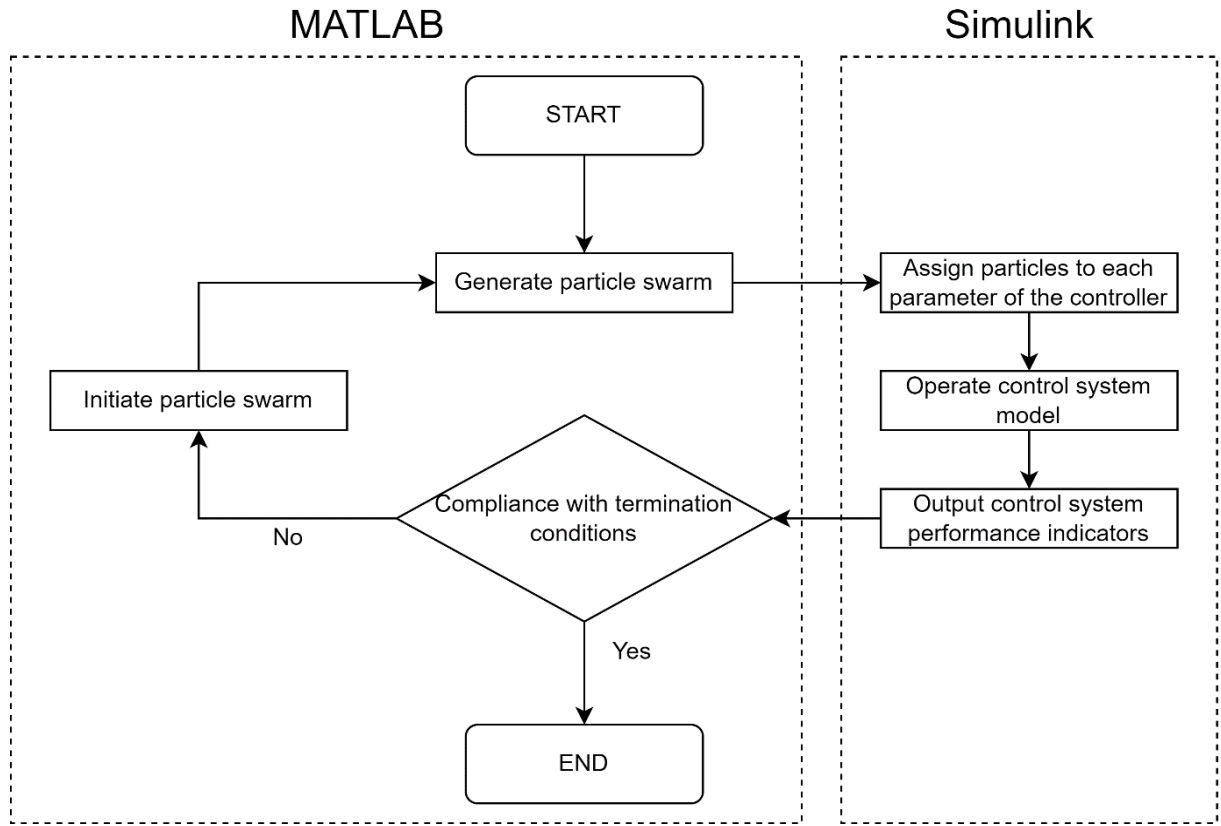

*Figure II-2. Process block diagram of controller parameters tuning by PSO.*

# III. Experimentation

In this chapter, we present the experimentation setup to validate our proposed fuzzy adaptive control system. We use a BlueROV2 heavy model (BlueRobotics 2024) in simulation built in Simulink. The ROV has 8 thrusters, patterned as shown in Fig. III-1, of which 4 are vectored actuators and 4 are vertical actuators, offering the vehicle full actuations in all 6 DOF. It weighs 12 kg and has a size of $547 \times 338 \times 254$ mm. The vehicle is designed for shallow water operations with a maximum rated depth of 100 m, which makes it a good candidate for under-ice exploration in this project.

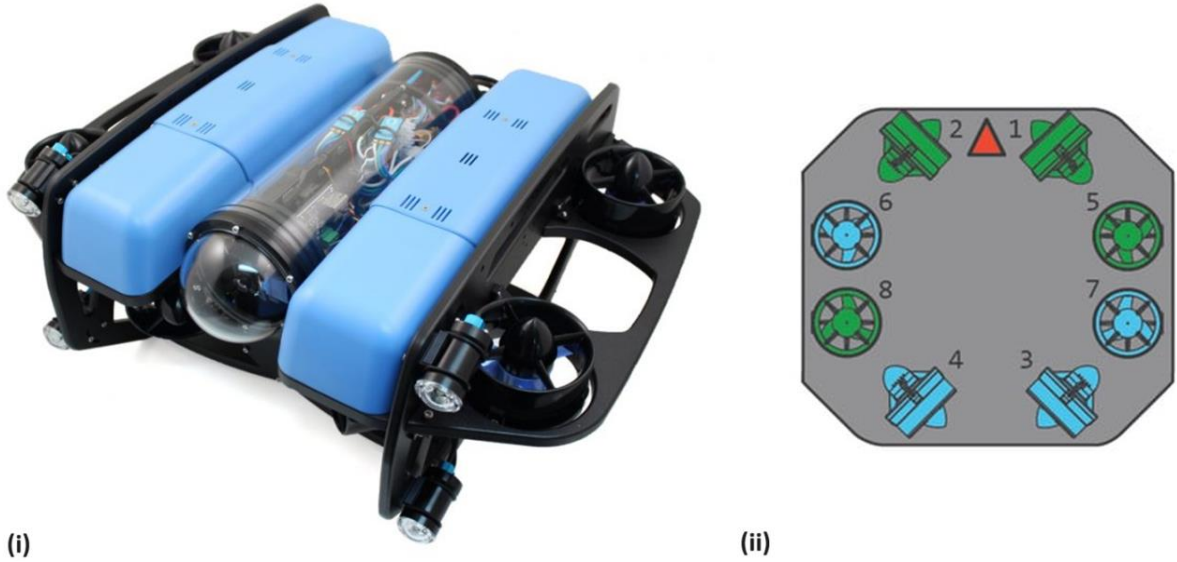

(i)　　　　　　　　　　　　　　　　　　　(ii)

*Figure III-1. BlueROV2 heavy with its thruster patterns.*

The system parameters of BlueROV2 heavy model have been well recognized (von Benzon 2022) and are listed in Appendix II. A Simulink model is built as shown in Fig. III-2. For the reference trajectory, a straight-line path is defined as

$$\eta_\mathrm{d} = \left[0.2t, 0.2t, 0, 0, 0, \frac{\pi}{4}\right]^T \tag{3.1}$$

in which $t$ is the simulation time. The ROV is required to travel continuously in north-east direction with a constant heading angle of 45 degrees while maintaining a constant moving speed and constant attitudes in depth, roll, and pitch. To simulate the bounded, low frequency disturbance, a constant disturbance as follows is added in the simulation but remains unknown to the control system.

$$\tau_d = [-1, 1, 2, 0.1, 0.1, 0]^T \tag{3.2}$$

The optimal controller parameters found by PSO algorithm, shown in Eqs. 4.1 and 4.2, are applied in the simulation.

For the fuzzy-based disturbance adaptation, 4 rules are utilized for each DOF. An example of the Gaussian membership functions used for body frame $q$ axis estimation is shown in Fig. III-3. The detailed rules and the corresponding Gaussian membership functions are listed in Appendix III.

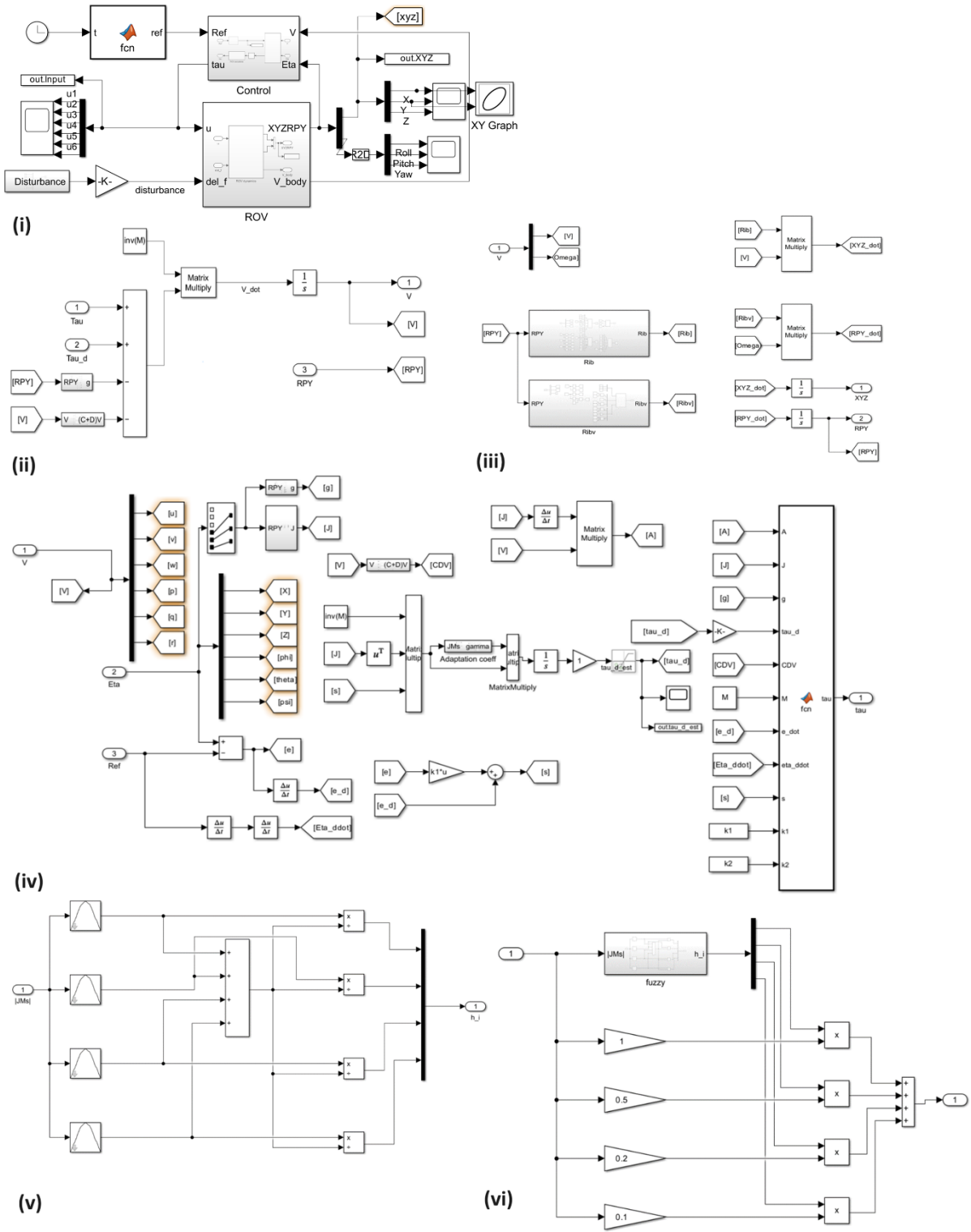

*Figure III-2. Simulink model used for ROV simulations. (i) ROV system, (ii) ROV dynamic system in body frame, (iii) ROV kinematic system in global frame, (iv) ROV fuzzy adaptive control system, (v) fuzzy system membership function example, (vi) fuzzy system defuzzification example.*

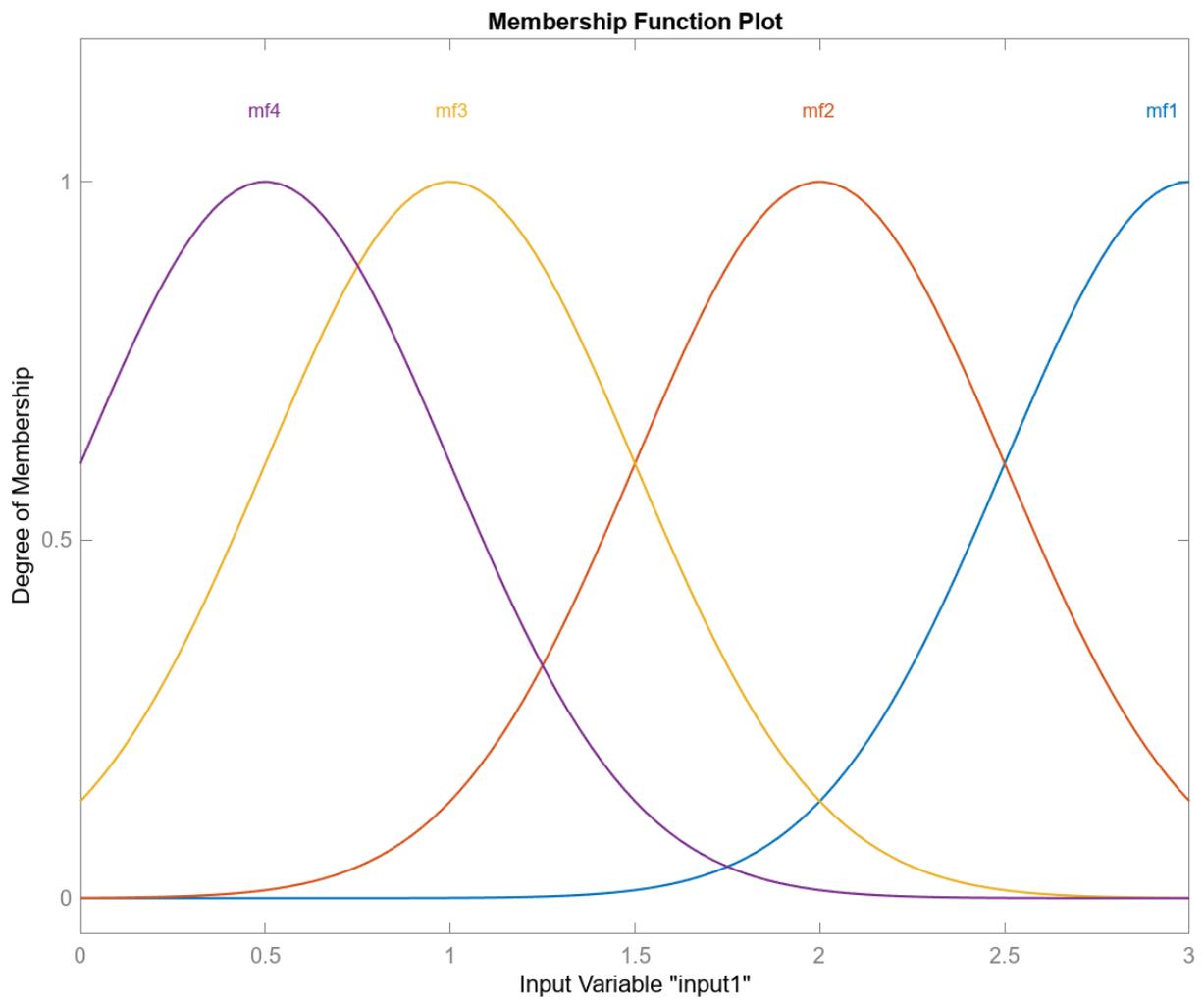

*Figure III-3. Membership functions used for disturbance estimation on q- direction in body frame.*

## IV. Results

In this chapter, we present the experiment results following the setup in Chapter 3. First, the PSO-based controller tuning results are presented. Using the fine-tunned controller parameters, the performances of a baseline backstepping controller that does not consider disturbances and the proposed fuzzy adaptive controller are compared based on the tracking errors along a straight-line reference trajectory under the influence of a constant environmental disturbance added in the simulations. Then, the disturbance estimation results by a constant adaptation and the proposed fuzzy varying adaptation are compared based on the estimation errors. Finally, the proposed fuzzy adaptive controller is applied on an ROV to follow a more complex trajectory in a real-life scenario and the simulation results are discussed.

### i. Controller parameter optimization results

To tune the 12 controller parameters, 100 particles are used in PSO for 100 iterations. The two cost function parameter matrices, $Q$ and $R$, are assigned to be identity matrices in this work. As shown in the process block diagram in Fig. II-2, in each iteration, the particle positions are assigned to each controller parameter with which the simulation of the control system is proceeded. The performance of the PSO-generated controller parameters is evaluated by the cost function defined in Eq. 2.32. The best cost is recorded after each iteration as shown in Fig. IV-1(i). The optimization results show that the controller parameters come to a convergence after approximately 100 iterations to achieve lower cost function results. According to the controller tuning results in Fig. IV-1(ii), the optimal controller parameters sets are

$$k_1 = diag(10, 1, 5.9, 1.7, 5.8, 0.8) \tag{4.1}$$

$$k_2 = diag(5.2, 10, 1, 5.8, 1.9, 5.5) \tag{4.2}$$

The controller parameters in Eqs. 4.1 and 4.2 are used in all simulations in this work.

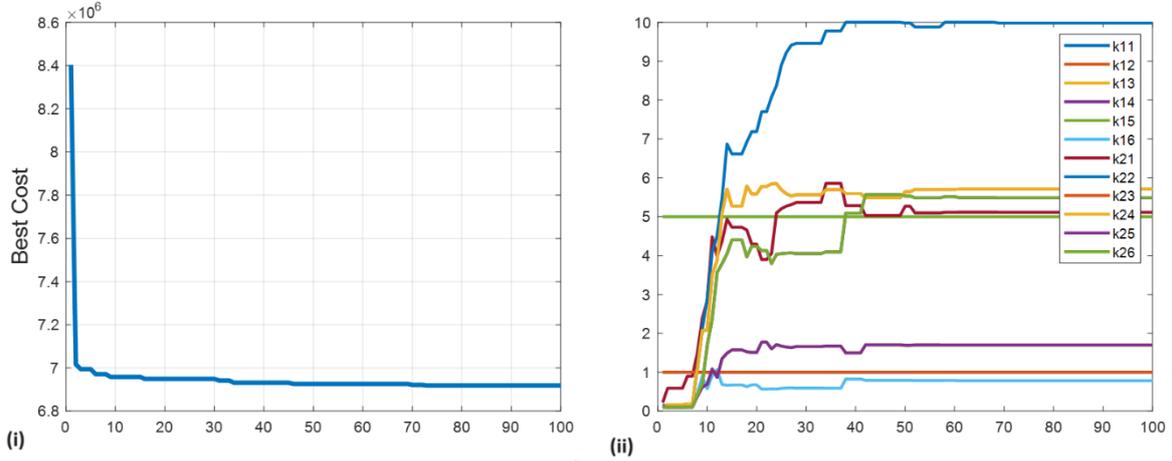

*Figure IV-1. Controller parameter tuning based on PSO method. (i) Best cost function values after each iteration; (ii) controller parameter convergence after each iteration.*

### ii.   Controller performance results

Utilizing the controller parameters determined by PSO method, we compare the performance of the ROV tracking the straight-line path by the baseline backstepping controller and the proposed fuzzy adaptive controller. Fig. IV-2 shows the system output by the baseline backstepping controller which does not consider the environmental disturbance. The influence of the constant disturbance on the ROV's path following is significant as the depth and all orientations are drifting from the reference target as shown. Fig. IV-3 shows the system output by the proposed fuzzy adaptive controller which estimates the environment disturbance. The trajectory tracking performance in this case is improved as the outputs in all 6 DOF are following the reference target. Note that even though in the plots of depth ($Z$) and pitch ($\theta$) tracking it shows large oscillations, the values (e.g. 2 cm in $Z$) are small comparing to the size of the ROV (25.4 cm in height), which means that such oscillations are negligible to evaluate the controller performance.

The performance comparison of the two controllers is clearer in the 3D views in Fig. IV-4. As shown, ROV with baseline controller is able to follow the approximate direction of the reference trajectory but can never track the trajectory due to the position drifting produced by environmental disturbance. In contrast, ROV with the proposed controller can follow the reference path from early stage of the task.

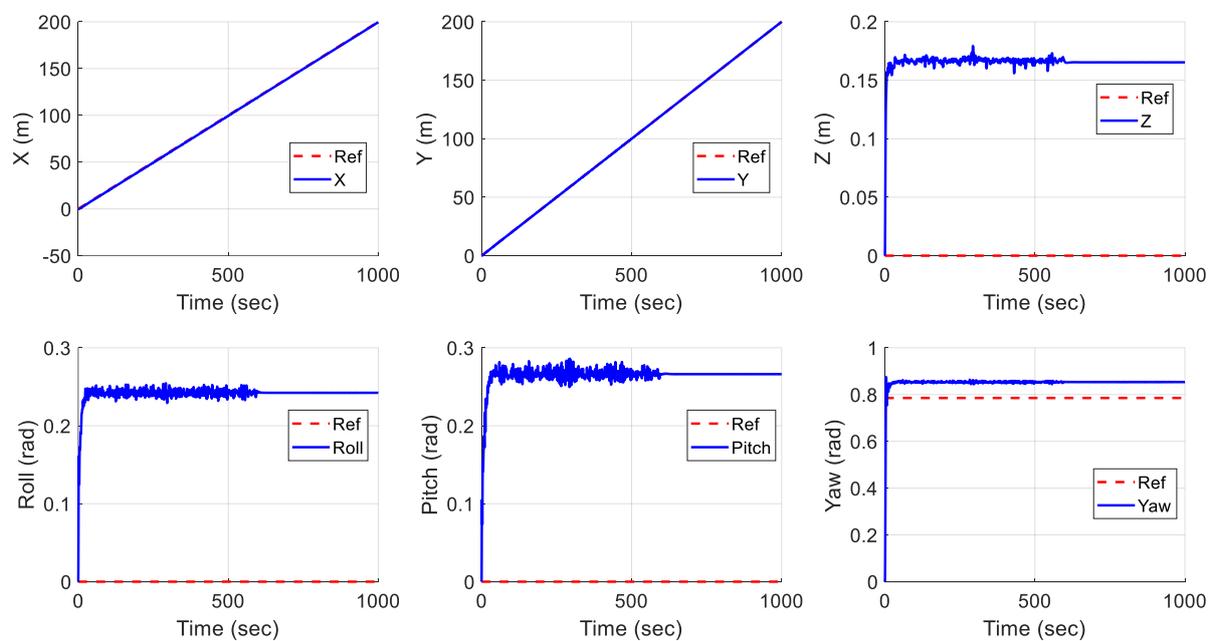

*Figure IV-2. Baseline backstepping controller trajectory tracking in disturbance, illustrated for all 6 DOF.*

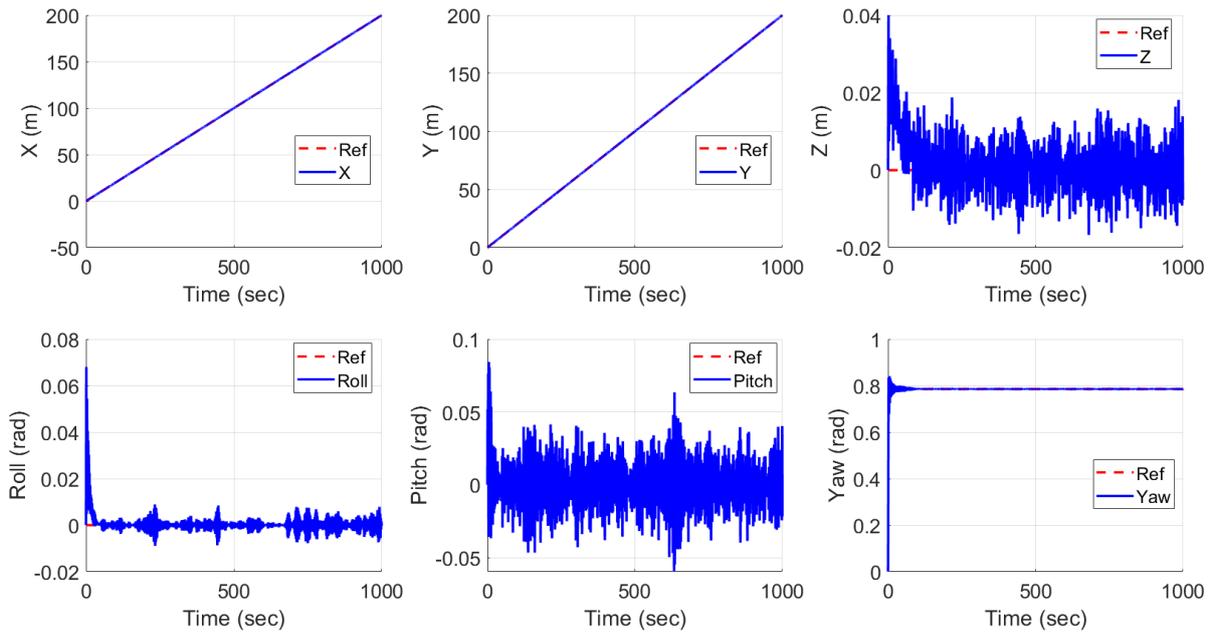

*Figure IV-3. Proposed fuzzy adaptive controller trajectory tracking in disturbance, illustrated for all 6 DOF.*

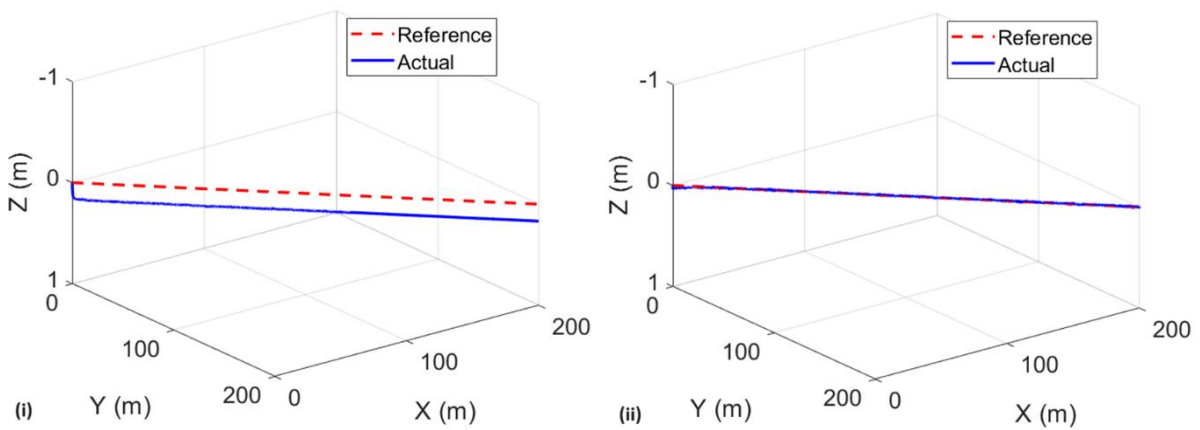

*Figure IV-4. 3D views of trajectory tracking results by (i) baseline backstepping controller, and (ii) proposed fuzzy adaptive controller.*

### iii.    Disturbance adaptation results

The efficacy of the proposed fuzzy adaptation law is examined by comparing its estimation results to those produced by constant adaptation rate coefficients. As shown in Fig. IV-5, even though theoretically the Lyapunov functions of the adaptation law guarantees stability, converge in estimation may not be successful in finite time, and this leads to larger output errors, which then further enlarge the estimate errors.

This is why the disturbance estimation results do not show a sign of convergence for the rotational disturbances as plotted in the figure. The estimations of translational disturbances show convergence but do not reach the real value of the disturbances.

In contrast, as shown in Fig. IV-6, in all 6 DOF, the fuzzy-based adaptation law enables the system to estimate the environmental disturbances correctly. All the estimations show convergence toward the actual values of the real disturbances. Note that there still exist oscillations in the estimation results for both translational and rotational disturbances, but the estimation errors, comparing to those resulted from constant adaptation rate coefficients, are much lower, and as indicated by the control system results illustrated in Figs. IV-3 and IV-4, such small amount of estimation errors does not have significant influence on the ROV's trajectory tracking performance.

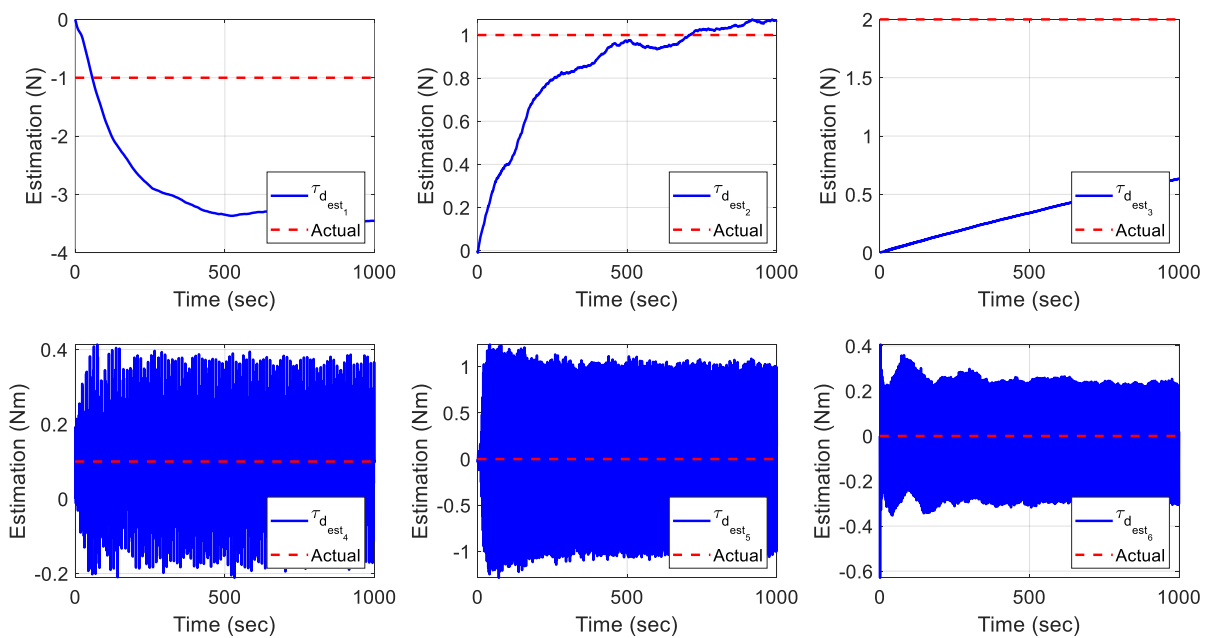

*Figure IV-5. Disturbance adaptation results by using constant adaptation rates.*

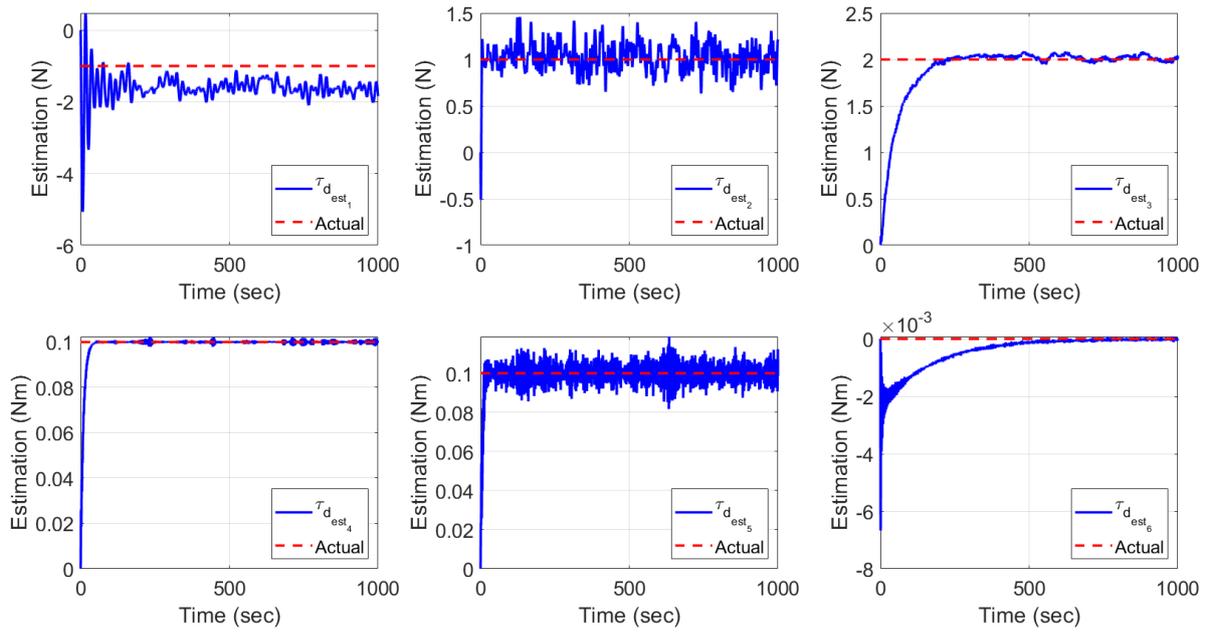

*Figure IV-6. Disturbance adaptation results by applying fuzzy-based varying adaptation rates.*

### iv. Demonstration of a real-life application

We further test the proposed adaptive control system on ROV in a real-life application scenario in which the vehicle is commanded to search in a $10 \times 10$ m pool, following a pre-planned square path sequence. As shown in Fig. IV-7, the vehicle moves in positive $x$- direction first with visible oscillations in the real-time trajectory but still follows the preset course, and after the first turn, the oscillation in its real-time trajectory decreases and the ROV is able to track the path smoothly afterward. The disturbance estimation results are shown in Fig. IV-8. At the beginning, these exist large estimation errors in the first three DOF, which explains the initial oscillations in the real-time trajectory, but the errors are gradually decreasing and eventually the disturbance estimation values converge to the real value in finite time, leading to a smoother trajectory tracking.

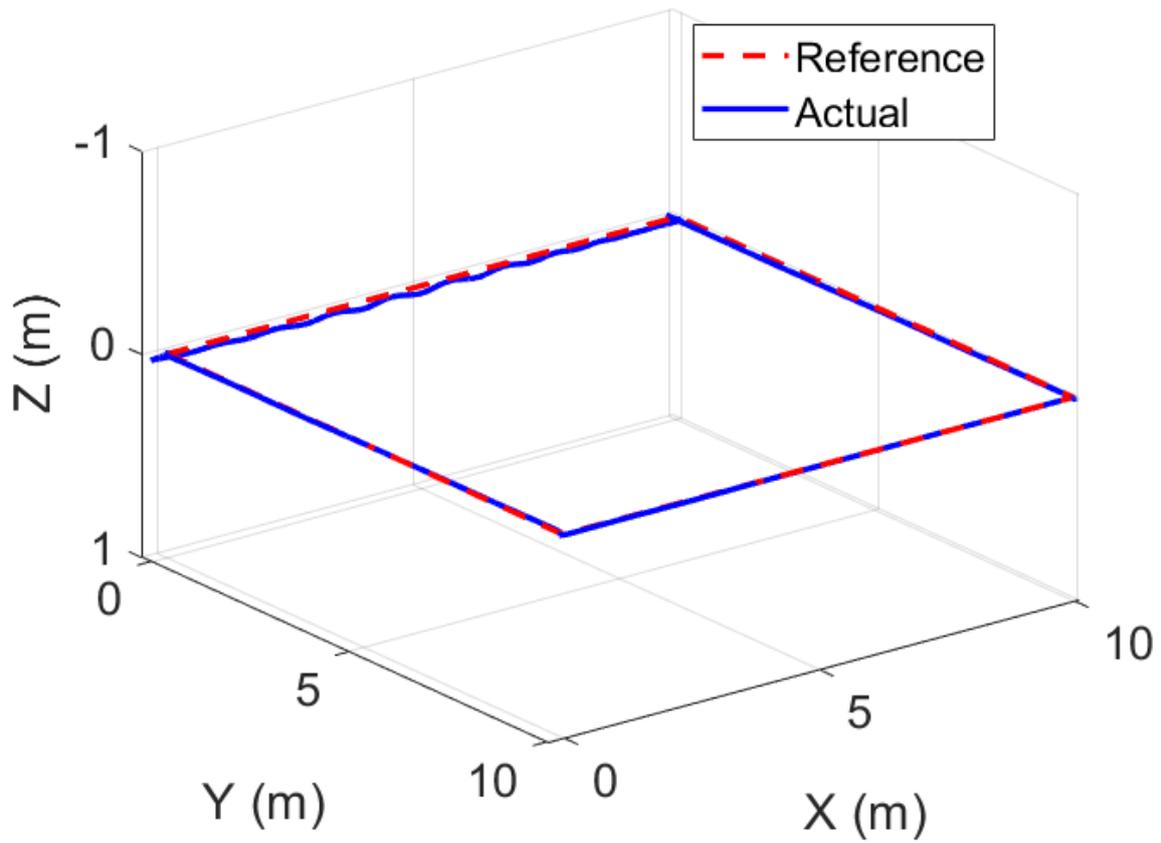

*Figure IV-7. 3D view of ROV following a square path sequence by utilizing the proposed fuzzy-based adaptive control system.*

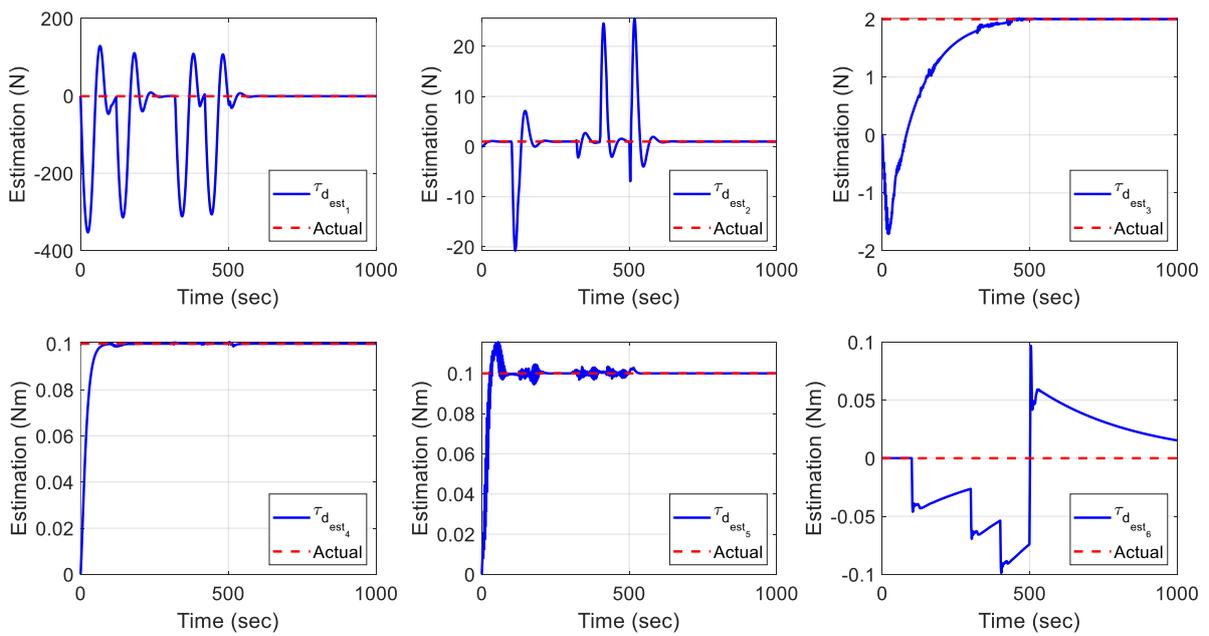

*Figure IV-8. Disturbance adaptation results of ROV following the square path sequence.*

# V.     Conclusion and future works

In conclusion, in this work, we designed a nonlinear control system with disturbance adaptation utilizing backstepping method based on Lyapunov functions to control a remotely operated vehicle (ROV) to follow a pre-planned trajectory. We applied particle swarm optimization (PSO) methodology to assist tuning the 12 controller parameters to achieve small tracking errors. We then implemented a type-1 Mamdani fuzzy inference system (FIS) to determine the varying adaptation rates based on the output errors to accomplish easier disturbance adaptation tuning and better disturbance estimation convergence. We also demonstrated the capability of tracking complex trajectory using the proposed control system on the ROV in a real-life application scenario. The simulation results have shown that the proposed methodology can help lower the influence of unknown environmental disturbances on the ROV motions and thereby enable the vehicle to perform more complicated underwater tasks in dynamic waters.

For future works, we would like to expand the proposed methodology so that it can be implemented on more universal ROV models. One problem with the universal implementation is that the proposed control system requires full-state feedback including the body frame velocities, which, in our case, are measured by a Doppler Velocity Logger (DVL) installed on the vehicle, but for the ROV models without such sensors on board, a full-state feedback control is not accessible, and therefore, we plan to modify the proposed control method to an output-feedback controller, which requires a design of an extended state observer that estimates both environmental disturbance and body frame velocities based on system output in global frame. In addition, to achieve better disturbance estimations, more fuzzy rules are to be added and a nonlinear adaptation law is needed.

## Appendix I: ROV parameter matrices

1. Inertia matrix

$$M = M_{rov} + M_a$$

$$M_{rov} = \begin{bmatrix} m & 0 & 0 & 0 & mz_g & -my_g \\ 0 & m & 0 & -mz_g & 0 & mx_g \\ 0 & 0 & m & my_g & -mx_g & 0 \\ 0 & -mz_g & my_g & I_x & -I_{xy} & -I_{xz} \\ mz_g & 0 & -mx_g & -I_{yx} & I_y & -I_{yz} \\ -my_g & mx_g & 0 & -I_{zx} & -I_{zy} & I_z \end{bmatrix}$$

$$M_a = diag(X_{\dot{u}}, Y_{\dot{v}}, Z_{\dot{w}}, K_{\dot{p}}, M_{\dot{q}}, N_{\dot{r}})$$

2. Coriolis matrix

$$C = C_{rov} + C_a$$

$$C_{rov} = \begin{bmatrix} 0 & 0 & 0 & 0 & mw & -mv \\ 0 & 0 & 0 & -mw & 0 & mu \\ 0 & 0 & 0 & mv & -mu & 0 \\ 0 & mw & -mv & 0 & I_z r & -I_z q \\ -mw & 0 & mu & -I_z r & 0 & I_x p \\ mv & -mu & 0 & I_y q & -I_x p & 0 \end{bmatrix}$$

$$C_a = \begin{bmatrix} 0 & 0 & 0 & 0 & -Z_{\dot{w}} w & Y_{\dot{v}} v \\ 0 & 0 & 0 & Z_{\dot{w}} w & 0 & -X_{\dot{u}} u \\ 0 & 0 & 0 & -Y_{\dot{v}} v & X_{\dot{u}} u & 0 \\ 0 & -Z_{\dot{w}} w & Y_{\dot{u}} u & 0 & -N_{\dot{r}} r & M_{\dot{q}} q \\ Z_{\dot{w}} w & 0 & -X_{\dot{u}} u & N_{\dot{r}} r & 0 & -K_{\dot{p}} p \\ -Y_{\dot{v}} v & X_{\dot{u}} u & 0 & -M_{\dot{q}} q & K_{\dot{p}} p & 0 \end{bmatrix}$$

3. Damping matrix

$$D = D_l + D_{nl}$$

$$D_l = diag(X_u, Y_v, Z_w, K_p, M_q, N_r)$$

$$D_{nl} = diag(X_{u|u|}|u|, Y_{v|v|}|v|, Z_{w|w|}|w|, K_{p|p|}|p|, M_{q|q|}|q|, N_{r|r|}|r|)$$

4. Gravity and buoyancy vector

$$g = \begin{bmatrix} (W-B)\sin\theta \\ -(W-B)\cos\theta\sin\phi \\ -(W-B)\cos\theta\cos\phi \\ -(y_g W - y_b B)\cos\theta\cos\phi + (z_g W - z_b B)\cos\theta\sin\phi \\ (z_g W - z_b B)\sin\theta + (x_g W - x_b B)\cos\theta\cos\phi \\ -(x_g W - x_b B)\cos\theta\sin\phi - (y_g W - y_b B)\sin\theta \end{bmatrix}$$

# Appendix II: BlueROV2 heavy system parameters

1. Inertia, gravity, and buoyancy

| Mass | $m$ | $13.5\ kg$ |
|---|---|---|
| Volume | $\nabla$ | $0.0135\ m^3$ |
| Moment of inertia | $I_x$ | $0.26\ kg \cdot m^2$ |
|  | $I_y$ | $0.23\ kg \cdot m^2$ |
|  | $I_z$ | $0.37\ kg \cdot m^2$ |
|  | $I_{xy}, I_{yz}, I_{zx}, I_{yx}, I_{zy}, I_{xz}$ | $0\ kg \cdot m^2$ |
| Center of gravity | $[x, y, z]_g$ | $[0,0,0]\ m$ |
| Center of buoyancy | $[x, y, z]_b$ | $[0,0,-0.01]\ m$ |

2. Coriolis

|  |  |  |
|---|---|---|
|  | $X_{\dot{u}}$ | $6.357\ kg$ |
|  | $Y_{\dot{v}}$ | $7.121\ kg$ |
|  | $Z_{\dot{w}}$ | $18.69\ kg$ |
|  | $K_{\dot{p}}$ | $0.1858\ kg$ |
|  | $M_{\dot{q}}$ | $0.1348\ kg$ |
|  | $N_{\dot{r}}$ | $0.2215\ kg$ |

3. Damping

|  |  |  |
|---|---|---|
|  | $X_u$ | $13.7\ N \cdot s/m$ |
|  | $Y_v$ | $0\ N \cdot s/m$ |
|  | $Z_w$ | $33\ N \cdot s/m$ |
|  | $K_p$ | $0\ N \cdot s/rad$ |
|  | $M_q$ | $0.8\ N \cdot s/rad$ |
|  | $N_r$ | $0 N \cdot s/rad$ |
|  | $X_{u|u|}$ | $141\ N \cdot s^2/m^2$ |
|  | $Y_{v|v|}$ | $217\ N \cdot s^2/m^2$ |
|  | $Z_{w|w|}$ | $190\ N \cdot s^2/m^2$ |
|  | $K_{p|p|}$ | $1.192\ N \cdot s^2/rad^2$ |
|  | $M_{q|q|}$ | $0.47\ N \cdot s^2/rad^2$ |
|  | $N_{r|r|}$ | $1.5\ N \cdot s^2/rad^2$ |

# Appendix III: Fuzzy rule base and membership functions

1. $u$-, $v$-, $w$- axes

Rules:

IF $|JM^{-1}s|_{1,2,3}$ is 5, THEN $\gamma_{1,2,3}$ is 100.

IF $|JM^{-1}s|_{1,2,3}$ is 2, THEN $\gamma_{1,2,3}$ is 50.

IF $|JM^{-1}s|_{1,2,3}$ is 1, THEN $\gamma_{1,2,3}$ is 20.

IF $|JM^{-1}s|_{1,2,3}$ is 0.5, THEN $\gamma_{1,2,3}$ is 10.

Membership functions:

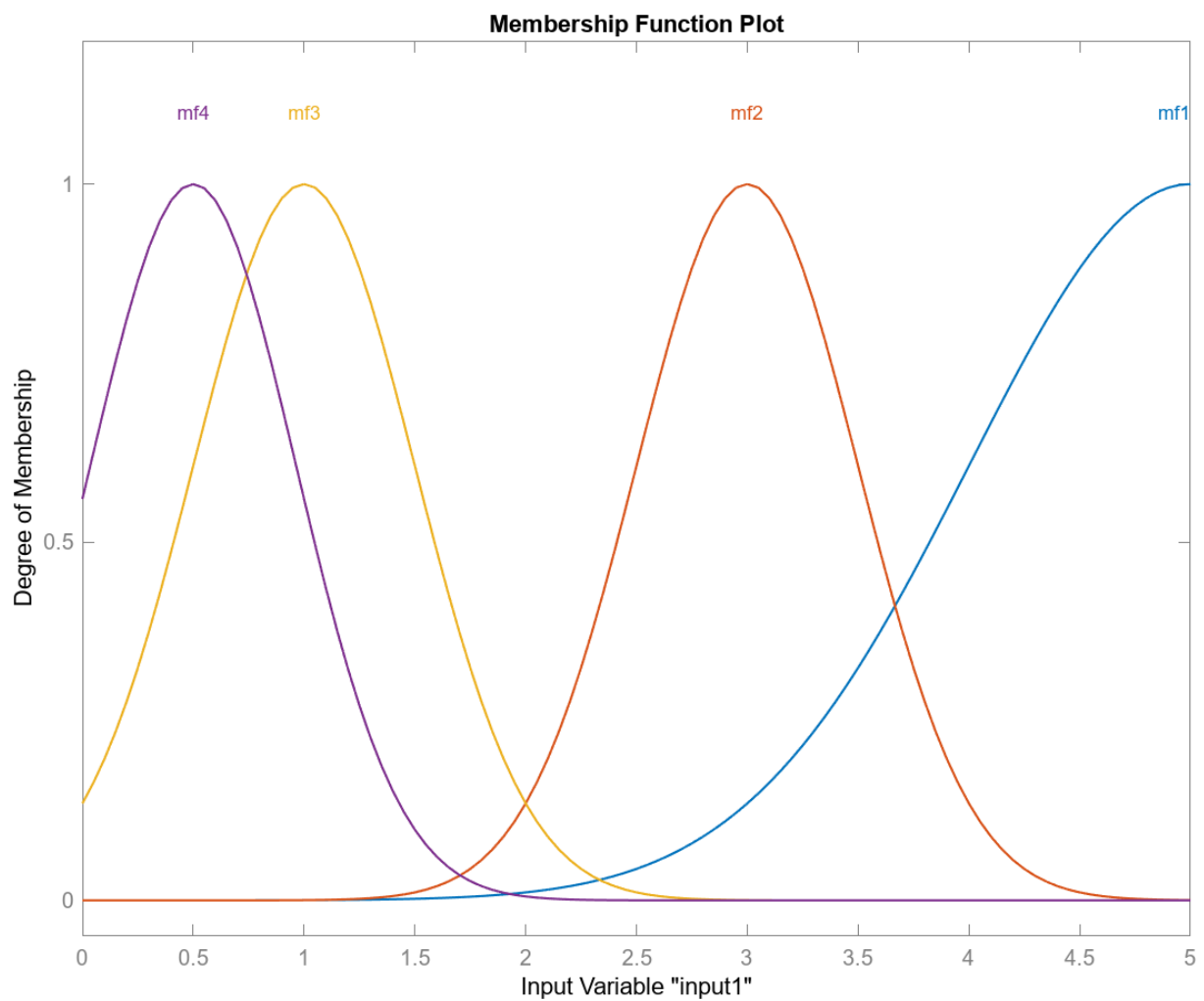

2. $p$-, $q$-, $r$- axes

Rules:

IF $|JM^{-1}s|_{4,5,6}$ is 3, THEN $\gamma_{4,5,6}$ is 1.

IF $|JM^{-1}s|_{4,5,6}$ is 2, THEN $\gamma_{4,5,6}$ is 0.5.

IF $|JM^{-1}s|_{4,5,6}$ is 1, THEN $\gamma_{4,5,6}$ is 0.2.

IF $|JM^{-1}s|_{4,5,6}$ is 0.5, THEN $\gamma_{4,5,6}$ is 0.1.

Membership functions:

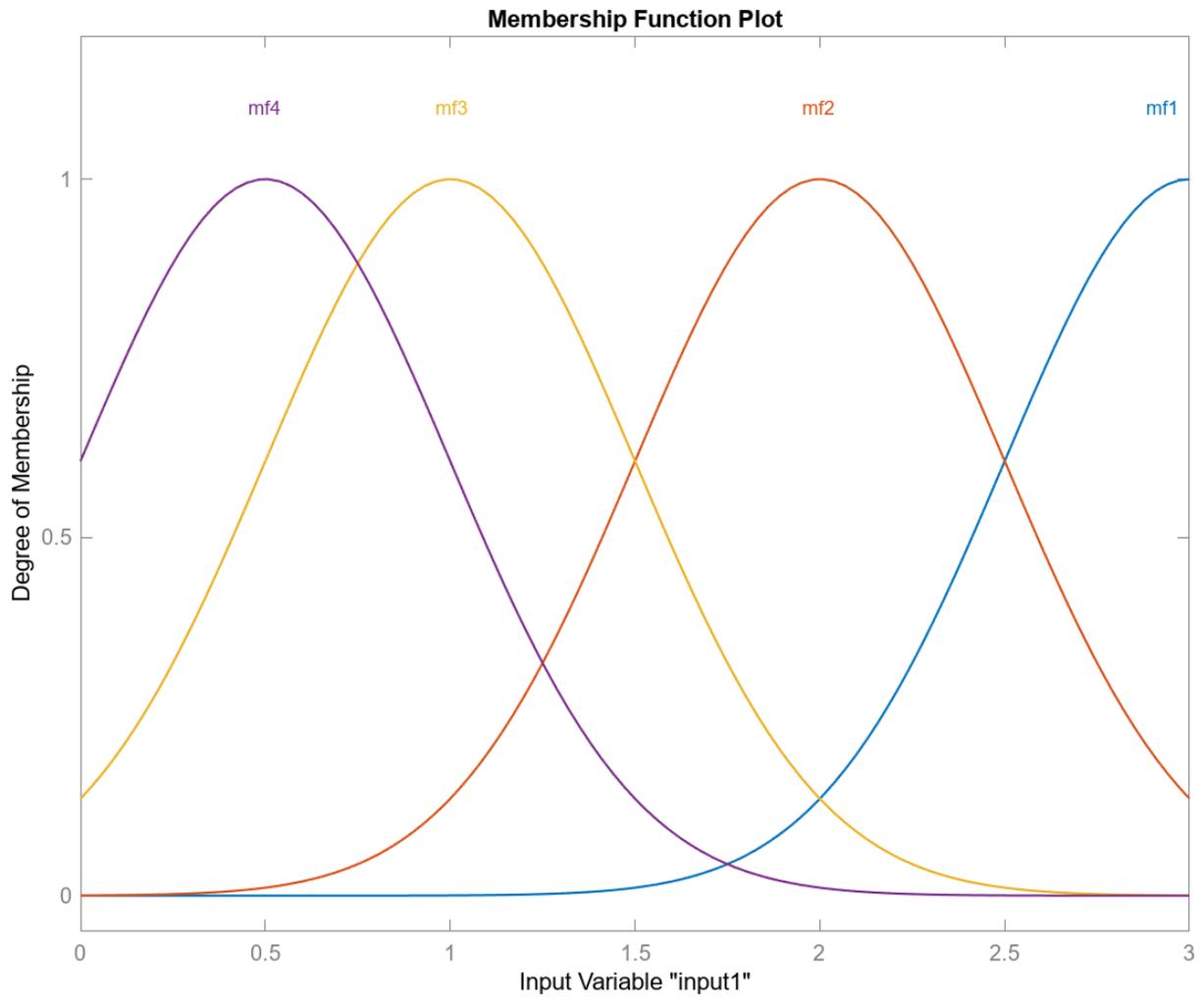